%% file: root.tex
\documentclass[letterpaper, 10 pt, conference]{ieeeconf}
\IEEEoverridecommandlockouts
\usepackage{amsmath,amssymb,amsfonts}
\usepackage{graphicx}
\usepackage{textcomp}
\usepackage{xcolor}
\def\BibTeX{{\rm B\kern-.05em{\sc i\kern-.025em b}\kern-.08em
    T\kern-.1667em\lower.7ex\hbox{E}\kern-.125emX}}

\usepackage{comment}

\usepackage{mathtools} %
\usepackage{booktabs} %
\usepackage{tikz} %

\usepackage[utf8]{inputenc} %
\usepackage{hyperref}       %
\usepackage{url}            %
\usepackage{amsfonts}       %
\usepackage{nicefrac}       %
\usepackage{microtype}      %
\usepackage{tabularx}
\usepackage{subcaption}
\usepackage{hyperref}
\usepackage{ragged2e}
\usepackage[noend]{algpseudocode}
\usepackage{algorithm}%
\usepackage{multicol}
\usepackage{gensymb}
\usepackage{wrapfig}

\input{vmr-symbols-vecbold}

\input{standard-macros}

\input{defs}

\newcommand{\fakeparagraph}[1]{\noindent {\bf #1 }}

\newcommand{\pluseq}{\mathrel{{+}{=}}}

\usepackage[capitalise,nameinlink]{cleveref}

\begin{document}

\title{Multi-Agent Active Search using Detection and Location Uncertainty
}
\author{Arundhati Banerjee$^1$, Ramina Ghods$^1$, Jeff Schneider$^1$
\thanks{$^{1}$A. Banerjee, R. Ghods and J. Schneider are with the School of Computer Science, Carnegie Mellon University, 
	Pittsburgh, PA 15213.
	{\tt\small \{arundhat, rghods, schneide\}@cs.cmu.edu} }
}

\maketitle

\begin{abstract}
Active search, in applications like environment monitoring or disaster response missions, involves autonomous agents detecting targets in a search space using decision making algorithms that adapt to the history of their observations. 
Active search algorithms must contend with two types of uncertainty: detection uncertainty and location uncertainty.
The more common approach in robotics is to focus on location uncertainty and remove detection uncertainty by thresholding the detection probability to zero or one.
In contrast, it is common in the sparse signal processing literature to assume the target location is accurate and instead focus on the uncertainty of its detection. 
In this work, we first propose an inference method to jointly handle both target detection and location uncertainty.  We then build a decision making algorithm on this inference method that uses Thompson sampling to enable  
decentralized multi-agent active search. 
We perform simulation experiments 
to show that our  
algorithms outperform competing baselines that only account for either target detection or location uncertainty.
We finally demonstrate the real world transferability of our algorithms using a realistic simulation environment we created on the Unreal Engine 4 platform with an AirSim plugin.  
\end{abstract}

\input{Introduction.tex}

\input{ProblemSetup.tex}

\input{RelatedWorks.tex}

\input{Algorithm.tex}
\input{Results.tex}

\clearpage
\bibliographystyle{IEEEtran}
\bibliography{IEEEabrv,ref}

\clearpage

\end{document}

%% file: vmr-symbols-vecbold.tex
\usepackage{amssymb}
\usepackage{amsfonts}
\usepackage{mathrsfs}
\usepackage{xspace}
\usepackage{bm}
\usepackage{upgreek}

\newcommand{\safemath}[2]{\newcommand{#1}{\ensuremath{#2}\xspace}}

\safemath{\bma}{\mathbf{a}}
\safemath{\bmb}{\mathbf{b}}
\safemath{\bmc}{\mathbf{c}}
\safemath{\bmd}{\mathbf{d}}
\safemath{\bme}{\mathbf{e}}
\safemath{\bmf}{\mathbf{f}}
\safemath{\bmg}{\mathbf{g}}
\safemath{\bmh}{\mathbf{h}}
\safemath{\bmi}{\mathbf{i}}
\safemath{\bmj}{\mathbf{j}}
\safemath{\bmk}{\mathbf{k}}
\safemath{\bml}{\mathbf{l}}
\safemath{\bmm}{\mathbf{m}}
\safemath{\bmn}{\mathbf{n}}
\safemath{\bmo}{\mathbf{o}}
\safemath{\bmp}{\mathbf{p}}
\safemath{\bmq}{\mathbf{q}}
\safemath{\bmr}{\mathbf{r}}
\safemath{\bms}{\mathbf{s}}
\safemath{\bmt}{\mathbf{t}}
\safemath{\bmu}{\mathbf{u}}
\safemath{\bmv}{\mathbf{v}}
\safemath{\bmw}{\mathbf{w}}
\safemath{\bmx}{\mathbf{x}}
\safemath{\bmy}{\mathbf{y}}
\safemath{\bmz}{\mathbf{z}}
\safemath{\bmzero}{\mathbf{0}}
\safemath{\bmone}{\mathbf{1}}

\bmdefine{\biad}{a}
\bmdefine{\bibd}{b}
\bmdefine{\bicd}{c}
\bmdefine{\bidd}{d}
\bmdefine{\bied}{e}
\bmdefine{\bifd}{f}
\bmdefine{\bigd}{g}
\bmdefine{\bihd}{h}
\bmdefine{\biid}{i}
\bmdefine{\bijd}{j}
\bmdefine{\bikd}{k}
\bmdefine{\bild}{l}
\bmdefine{\bimd}{m}
\bmdefine{\bind}{n}
\bmdefine{\biod}{o}
\bmdefine{\bipd}{p}
\bmdefine{\biqd}{q}
\bmdefine{\bird}{r}
\bmdefine{\bisd}{s}
\bmdefine{\bitd}{t}
\bmdefine{\biud}{u}
\bmdefine{\bivd}{v}
\bmdefine{\biwd}{w}
\bmdefine{\bixd}{x}
\bmdefine{\biyd}{y}
\bmdefine{\bizd}{z}

\bmdefine{\bixid}{\xi}
\bmdefine{\bilambdad}{\lambda}
\bmdefine{\bimud}{\mu}
\bmdefine{\bithetad}{\theta}
\bmdefine{\biphid}{\phi}
\bmdefine{\bideltad}{\delta}

\safemath{\bmia}{\biad}
\safemath{\bmib}{\bibd}
\safemath{\bmic}{\bicd}
\safemath{\bmid}{\bidd}
\safemath{\bmie}{\bied}
\safemath{\bmif}{\bifd}
\safemath{\bmig}{\bigd}
\safemath{\bmih}{\bihd}
\safemath{\bmii}{\biid}
\safemath{\bmij}{\bijd}
\safemath{\bmik}{\bikd}
\safemath{\bmil}{\bild}
\safemath{\bmim}{\bimd}
\safemath{\bmin}{\bind}
\safemath{\bmio}{\biod}
\safemath{\bmip}{\bipd}
\safemath{\bmiq}{\biqd}
\safemath{\bmir}{\bird}
\safemath{\bmis}{\bisd}
\safemath{\bmit}{\bitd}
\safemath{\bmiu}{\biud}
\safemath{\bmiv}{\bivd}
\safemath{\bmiw}{\biwd}
\safemath{\bmix}{\bixd}
\safemath{\bmiy}{\biyd}
\safemath{\bmiz}{\bizd}

\safemath{\bmxi}{\bixid}
\safemath{\bmlambda}{\bilambdad}
\safemath{\bmmu}{\bimud}
\safemath{\bmtheta}{\bithetad}
\safemath{\bmphi}{\biphid}
\safemath{\bmdelta}{\bideltad}

\safemath{\bA}{\mathbf{A}}
\safemath{\bB}{\mathbf{B}}
\safemath{\bC}{\mathbf{C}}
\safemath{\bD}{\mathbf{D}}
\safemath{\bE}{\mathbf{E}}
\safemath{\bF}{\mathbf{F}}
\safemath{\bG}{\mathbf{G}}
\safemath{\bH}{\mathbf{H}}
\safemath{\bI}{\mathbf{I}}
\safemath{\bJ}{\mathbf{J}}
\safemath{\bK}{\mathbf{K}}
\safemath{\bL}{\mathbf{L}}
\safemath{\bM}{\mathbf{M}}
\safemath{\bN}{\mathbf{N}}
\safemath{\bO}{\mathbf{O}}
\safemath{\bP}{\mathbf{P}}
\safemath{\bQ}{\mathbf{Q}}
\safemath{\bR}{\mathbf{R}}
\safemath{\bS}{\mathbf{S}}
\safemath{\bT}{\mathbf{T}}
\safemath{\bU}{\mathbf{U}}
\safemath{\bV}{\mathbf{V}}
\safemath{\bW}{\mathbf{W}}
\safemath{\bX}{\mathbf{X}}
\safemath{\bY}{\mathbf{Y}}
\safemath{\bZ}{\mathbf{Z}}

\safemath{\bZero}{\mathbf{0}}
\safemath{\bOne}{\mathbf{1}}
\safemath{\bDelta}{\mathbf{\Delta}}
\safemath{\bLambda}{\boldsymbol\Lambda}
\safemath{\bPhi}{\mathbf{\Upphi}}
\safemath{\bSigma}{\mathbf{\Upsigma}}
\safemath{\bOmega}{\mathbf{\Upomega}}
\safemath{\bTheta}{\mathbf{\Uptheta}}

\bmdefine{\biAd}{A}
\bmdefine{\biBd}{B}
\bmdefine{\biCd}{C}
\bmdefine{\biDd}{D}
\bmdefine{\biEd}{E}
\bmdefine{\biFd}{F}
\bmdefine{\biGd}{G}
\bmdefine{\biHd}{H}
\bmdefine{\biId}{I}
\bmdefine{\biJd}{J}
\bmdefine{\biKd}{K}
\bmdefine{\biLd}{L}
\bmdefine{\biMd}{M}
\bmdefine{\biOd}{N}
\bmdefine{\biPd}{O}
\bmdefine{\biQd}{P}
\bmdefine{\biRd}{R}
\bmdefine{\biSd}{S}
\bmdefine{\biTd}{T}
\bmdefine{\biUd}{U}
\bmdefine{\biVd}{V}
\bmdefine{\biWd}{W}
\bmdefine{\biXd}{X}
\bmdefine{\biYd}{Y}
\bmdefine{\biZd}{Z}

\bmdefine{\biDelta}{\Delta}
\bmdefine{\biLambda}{\Lambda}
\bmdefine{\biPhi}{\Phi}
\bmdefine{\biSigma}{\Sigma}
\bmdefine{\biOmega}{\Omega}
\bmdefine{\biTheta}{\Theta}

\safemath{\bimA}{\biAd}
\safemath{\bimB}{\biBd}
\safemath{\bimC}{\biCd}
\safemath{\bimD}{\biDd}
\safemath{\bimE}{\biEd}
\safemath{\bimF}{\biFd}
\safemath{\bimG}{\biGd}
\safemath{\bimH}{\biHd}
\safemath{\bimI}{\biId}
\safemath{\bimJ}{\biJd}
\safemath{\bimK}{\biKd}
\safemath{\bimL}{\biLd}
\safemath{\bimM}{\biMd}
\safemath{\bimN}{\biNd}
\safemath{\bimO}{\biOd}
\safemath{\bimP}{\biPd}
\safemath{\bimQ}{\biQd}
\safemath{\bimR}{\biRd}
\safemath{\bimS}{\biSd}
\safemath{\bimT}{\biTd}
\safemath{\bimU}{\biUd}
\safemath{\bimV}{\biVd}
\safemath{\bimW}{\biWd}
\safemath{\bimX}{\biXd}
\safemath{\bimY}{\biYd}
\safemath{\bimZ}{\biZd}

\safemath{\bimDelta}{\biDelta}
\safemath{\bimLambda}{\biLambda}
\safemath{\bimPhi}{\biPhi}
\safemath{\bimSigma}{\biSigma}
\safemath{\bimOmega}{\biOmega}
\safemath{\bimTheta}{\biTheta}

\safemath{\setA}{\mathcal{A}}
\safemath{\setB}{\mathcal{B}}
\safemath{\setC}{\mathcal{C}}
\safemath{\setD}{\mathcal{D}}
\safemath{\setE}{\mathcal{E}}
\safemath{\setF}{\mathcal{F}}
\safemath{\setG}{\mathcal{G}}
\safemath{\setH}{\mathcal{H}}
\safemath{\setI}{\mathcal{I}}
\safemath{\setJ}{\mathcal{J}}
\safemath{\setK}{\mathcal{K}}
\safemath{\setL}{\mathcal{L}}
\safemath{\setM}{\mathcal{M}}
\safemath{\setN}{\mathcal{N}}
\safemath{\setO}{\mathcal{O}}
\safemath{\setP}{\mathcal{P}}
\safemath{\setQ}{\mathcal{Q}}
\safemath{\setR}{\mathcal{R}}
\safemath{\setS}{\mathcal{S}}
\safemath{\setT}{\mathcal{T}}
\safemath{\setU}{\mathcal{U}}
\safemath{\setV}{\mathcal{V}}
\safemath{\setW}{\mathcal{W}}
\safemath{\setX}{\mathcal{X}}
\safemath{\setY}{\mathcal{Y}}
\safemath{\setZ}{\mathcal{Z}}
\safemath{\emptySet}{\varnothing}

\safemath{\colA}{\mathscr{A}}
\safemath{\colB}{\mathscr{B}}
\safemath{\colC}{\mathscr{C}}
\safemath{\colD}{\mathscr{D}}
\safemath{\colE}{\mathscr{E}}
\safemath{\colF}{\mathscr{F}}
\safemath{\colG}{\mathscr{G}}
\safemath{\colH}{\mathscr{H}}
\safemath{\colI}{\mathscr{I}}
\safemath{\colJ}{\mathscr{J}}
\safemath{\colK}{\mathscr{K}}
\safemath{\colL}{\mathscr{L}}
\safemath{\colM}{\mathscr{M}}
\safemath{\colN}{\mathscr{N}}
\safemath{\colO}{\mathscr{O}}
\safemath{\colP}{\mathscr{P}}
\safemath{\colQ}{\mathscr{Q}}
\safemath{\colR}{\mathscr{R}}
\safemath{\colS}{\mathscr{S}}
\safemath{\colT}{\mathscr{T}}
\safemath{\colU}{\mathscr{U}}
\safemath{\colV}{\mathscr{V}}
\safemath{\colW}{\mathscr{W}}
\safemath{\colX}{\mathscr{X}}
\safemath{\colY}{\mathscr{Y}}
\safemath{\colZ}{\mathscr{Z}}

\safemath{\opA}{\mathbb{A}}
\safemath{\opB}{\mathbb{B}}
\safemath{\opC}{\mathbb{C}}
\safemath{\opD}{\mathbb{D}}
\safemath{\opE}{\mathbb{E}}
\safemath{\opF}{\mathbb{F}}
\safemath{\opG}{\mathbb{G}}
\safemath{\opH}{\mathbb{H}}
\safemath{\opI}{\mathbb{I}}
\safemath{\opJ}{\mathbb{J}}
\safemath{\opK}{\mathbb{K}}
\safemath{\opL}{\mathbb{L}}
\safemath{\opM}{\mathbb{M}}
\safemath{\opN}{\mathbb{N}}
\safemath{\opO}{\mathbb{O}}
\safemath{\opP}{\mathbb{P}}
\safemath{\opQ}{\mathbb{Q}}
\safemath{\opR}{\mathbb{R}}
\safemath{\opS}{\mathbb{S}}
\safemath{\opT}{\mathbb{T}}
\safemath{\opU}{\mathbb{U}}
\safemath{\opV}{\mathbb{V}}
\safemath{\opW}{\mathbb{W}}
\safemath{\opX}{\mathbb{X}}
\safemath{\opY}{\mathbb{Y}}
\safemath{\opZ}{\mathbb{Z}}
\safemath{\opZero}{\mathbb{O}}
\safemath{\identityop}{\opI}

\safemath{\veca}{\bma}
\safemath{\vecb}{\bmb}
\safemath{\vecc}{\bmc}
\safemath{\vecd}{\bmd}
\safemath{\vece}{\bme}
\safemath{\vecf}{\bmf}
\safemath{\vecg}{\bmg}
\safemath{\vech}{\bmh}
\safemath{\veci}{\bmi}
\safemath{\vecj}{\bmj}
\safemath{\veck}{\bmk}
\safemath{\vecl}{\bml}
\safemath{\vecm}{\bmm}
\safemath{\vecn}{\bmn}
\safemath{\veco}{\bmo}
\safemath{\vecp}{\bmp}
\safemath{\vecq}{\bmq}
\safemath{\vecr}{\bmr}
\safemath{\vecs}{\bms}
\safemath{\vect}{\bmt}
\safemath{\vecu}{\bmu}
\safemath{\vecv}{\bmv}
\safemath{\vecw}{\bmw}
\safemath{\vecx}{\bmx}
\safemath{\vecy}{\bmy}
\safemath{\vecz}{\bmz}

\safemath{\veczero}{\bmzero}
\safemath{\vecone}{\bmone}
\safemath{\vecxi}{\bmxi}
\safemath{\veclambda}{\bmlambda}
\safemath{\vecmu}{\bmmu}
\safemath{\vectheta}{\bmtheta}
\safemath{\vecphi}{\bmphi}
\safemath{\vecdelta}{\bmdelta}

\safemath{\matA}{\bA}
\safemath{\matB}{\bB}
\safemath{\matC}{\bC}
\safemath{\matD}{\bD}
\safemath{\matE}{\bE}
\safemath{\matF}{\bF}
\safemath{\matG}{\bG}
\safemath{\matH}{\bH}
\safemath{\matI}{\bI}
\safemath{\matJ}{\bJ}
\safemath{\matK}{\bK}
\safemath{\matL}{\bL}
\safemath{\matM}{\bM}
\safemath{\matN}{\bN}
\safemath{\matO}{\bO}
\safemath{\matP}{\bP}
\safemath{\matQ}{\bQ}
\safemath{\matR}{\bR}
\safemath{\matS}{\bS}
\safemath{\matT}{\bT}
\safemath{\matU}{\bU}
\safemath{\matV}{\bV}
\safemath{\matW}{\bW}
\safemath{\matX}{\bX}
\safemath{\matY}{\bY}
\safemath{\matZ}{\bZ}
\safemath{\matzero}{\bmzero}

\safemath{\matDelta}{\bDelta}
\safemath{\matLambda}{\bLambda}
\safemath{\matPhi}{\bPhi}
\safemath{\matSigma}{\bSigma}
\safemath{\matOmega}{\bOmega}
\safemath{\matTheta}{\bTheta}

\safemath{\matidentity}{\matI}
\safemath{\matone}{\matO}

\safemath{\rnda}{A}
\safemath{\rndb}{B}
\safemath{\rndc}{C}
\safemath{\rndd}{D}
\safemath{\rnde}{E}
\safemath{\rndf}{F}
\safemath{\rndg}{G}
\safemath{\rndh}{H}
\safemath{\rndi}{I}
\safemath{\rndj}{J}
\safemath{\rndk}{K}
\safemath{\rndl}{L}
\safemath{\rndm}{M}
\safemath{\rndn}{N}
\safemath{\rndo}{O}
\safemath{\rndp}{P}
\safemath{\rndq}{Q}
\safemath{\rndr}{R}
\safemath{\rnds}{S}
\safemath{\rndt}{T}
\safemath{\rndu}{U}
\safemath{\rndv}{V}
\safemath{\rndw}{W}
\safemath{\rndx}{X}
\safemath{\rndy}{Y}
\safemath{\rndz}{Z}

\safemath{\rveca}{\bimA}
\safemath{\rvecb}{\bimB}
\safemath{\rvecc}{\bimC}
\safemath{\rvecd}{\bimD}
\safemath{\rvece}{\bimE}
\safemath{\rvecf}{\bimF}
\safemath{\rvecg}{\bimG}
\safemath{\rvech}{\bimH}
\safemath{\rveci}{\bimI}
\safemath{\rvecj}{\bimJ}
\safemath{\rveck}{\bimK}
\safemath{\rvecl}{\bimL}
\safemath{\rvecm}{\bimM}
\safemath{\rvecn}{\bimN}
\safemath{\rveco}{\bomO}
\safemath{\rvecp}{\bimP}
\safemath{\rvecq}{\bimQ}
\safemath{\rvecr}{\bimR}
\safemath{\rvecs}{\bimS}
\safemath{\rvect}{\bimT}
\safemath{\rvecu}{\bimU}
\safemath{\rvecv}{\bimV}
\safemath{\rvecw}{\bimW}
\safemath{\rvecx}{\bimX}
\safemath{\rvecy}{\bimY}
\safemath{\rvecz}{\bimZ}

\safemath{\rvecxi}{\bmxi}
\safemath{\rveclambda}{\bmlambda}
\safemath{\rvecmu}{\bmmu}
\safemath{\rvectheta}{\bmtheta}
\safemath{\rvecphi}{\bmphi}

\safemath{\rmatA}{\bimA}
\safemath{\rmatB}{\bimB}
\safemath{\rmatC}{\bimC}
\safemath{\rmatD}{\bimD}
\safemath{\rmatE}{\bimE}
\safemath{\rmatF}{\bimF}
\safemath{\rmatG}{\bimG}
\safemath{\rmatH}{\bimH}
\safemath{\rmatI}{\bimI}
\safemath{\rmatJ}{\bimJ}
\safemath{\rmatK}{\bimK}
\safemath{\rmatL}{\bimL}
\safemath{\rmatM}{\bimM}
\safemath{\rmatN}{\bimN}
\safemath{\rmatO}{\bimO}
\safemath{\rmatP}{\bimP}
\safemath{\rmatQ}{\bimQ}
\safemath{\rmatR}{\bimR}
\safemath{\rmatS}{\bimS}
\safemath{\rmatT}{\bimT}
\safemath{\rmatU}{\bimU}
\safemath{\rmatV}{\bimV}
\safemath{\rmatW}{\bimW}
\safemath{\rmatX}{\bimX}
\safemath{\rmatY}{\bimY}
\safemath{\rmatZ}{\bimZ}

\safemath{\rmatDelta}{\bimDelta}
\safemath{\rmatLambda}{\bimLambda}
\safemath{\rmatPhi}{\bimPhi}
\safemath{\rmatSigma}{\bimSigma}
\safemath{\rmatOmega}{\bimOmega}
\safemath{\rmatTheta}{\bimTheta}

%% file: standard-macros.tex
\usepackage{amssymb}
\usepackage{amsfonts}
\usepackage{mathrsfs}
\usepackage{xspace}
\usepackage{bm}
\usepackage{fancyref}
\usepackage{textcomp}

\usepackage{multirow}
\usepackage{stmaryrd}

\newenvironment{textbmatrix}{	\setlength{\arraycolsep}{2.5pt}%
								\big[\begin{matrix}}{\end{matrix}\big]%
								\raisebox{0.08ex}{\vphantom{M}}}

\def\be{\begin{equation}}
\def\ee{\end{equation}}
\def\een{\nonumber \end{equation}}
\def\mat{\begin{bmatrix}}
\def\emat{\end{bmatrix}}
\def\btm{\begin{textbmatrix}}
\def\etm{\end{textbmatrix}}

\def\ba#1\ea{\begin{align}#1\end{align}}
\def\bas#1\eas{\begin{align*}#1\end{align*}}
\def\bs#1\es{\begin{split}#1\end{split}} 
\def\bg#1\eg{\begin{gather}#1\end{gather}}
\def\bml#1\eml{\begin{multline}#1\end{multline}}
\def\bi#1\ei{\begin{itemize}#1\end{itemize}}

\DeclareMathOperator*{\argmax}{arg\;max}		%

\safemath{\dirac}{\delta}					%
\safemath{\krond}{\dirac}					%
\safemath{\upto}{\uparrow}
\safemath{\downto}{\downarrow}
\safemath{\iu}{j}							%
\safemath{\ev}{\lambda}						%
\safemath{\hilseqspace}{l^{2}}				%
\newcommand{\banachfunspace}[1]{\setL^{#1}}	%
\safemath{\hilfunspace}{\banachfunspace{2}}	%

\safemath{\SNR}{\textsf{SNR}} 				%
\safemath{\PAR}{\textsf{PAR}} 				%
\safemath{\No}{N_0}							%
\safemath{\Es}{E_s}							%
\safemath{\Eb}{E_b}							%
\safemath{\EbNo}{\frac{\Eb}{\No}}
\safemath{\EsNo}{\frac{\Es}{\No}}

\DeclareMathOperator{\CHop}{\ensuremath{\opH}} %
\safemath{\tvir}{\rndh_{\CHop}}				%
\safemath{\tvtf}{\rndl_{\CHop}}				%
\safemath{\spf}{\rnds_{\CHop}}				%
\safemath{\bff}{H_{\CHop}}					%

\safemath{\ircf}{r_{h}}						%
\safemath{\tftvcf}{r_{s}}					%
\safemath{\tfcf}{r_{l}}						%
\safemath{\bfcf}{r_{H}}						%

\safemath{\tcorr}{c_h}						%
\safemath{\scf}{c_{s}}						%
\safemath{\tfcorr}{c_{l}}					%
\safemath{\fcorr}{c_{H}}						%

\safemath{\mi}{I}							%
\safemath{\capacity}{C}						%

\safemath{\normal}{\mathcal{N}}			%
\safemath{\jpg}{\mathcal{CN}}			%
\safemath{\mchain}{\leftrightarrow}		%

\safemath{\dB}{\,\mathrm{dB}}
\safemath{\dBm}{\,\mathrm{dBm}}
\safemath{\Hz}{\,\mathrm{Hz}}
\safemath{\kHz}{\,\mathrm{kHz}}
\safemath{\MHz}{\,\mathrm{MHz}}
\safemath{\GHz}{\,\mathrm{GHz}}
\safemath{\s}{\,\mathrm{s}}
\safemath{\ms}{\,\mathrm{ms}}
\safemath{\mus}{\,\mathrm{\text{\textmu}s}}
\safemath{\ns}{\,\mathrm{ns}}
\safemath{\ps}{\,\mathrm{ps}}
\safemath{\meter}{\,\mathrm{m}}
\safemath{\mm}{\,\mathrm{mm}}
\safemath{\cm}{\,\mathrm{cm}}
\safemath{\m}{\,\mathrm{m}}
\safemath{\W}{\,\mathrm{W}}
\safemath{\mW}{\, \mathrm{mW}}
\safemath{\J}{\,\mathrm{J}}
\safemath{\K}{\,\mathrm{K}}
\safemath{\bit}{\,\mathrm{bit}}
\safemath{\nat}{\,\mathrm{nat}}

\safemath{\define}{\triangleq}			%

\safemath{\equivalent}{\sim}
\safemath{\distas}{\sim}					%
\safemath{\sdiff}{\Delta}				%

\safemath{\reals}{\mathbb{R}}
\safemath{\positivereals}{\reals_{+}}
\safemath{\integers}{\mathbb{Z}}
\safemath{\posint}{\integers_{+}}
\safemath{\naturals}{\mathbb{N}}
\safemath{\posnaturals}{\naturals_{+}}
\safemath{\complexset}{\mathbb{C}}
\safemath{\rationals}{\mathbb{Q}}

\newcommand*{\fancyrefapplabelprefix}{app}		%
\newcommand*{\fancyrefthmlabelprefix}{thm}		%
\newcommand*{\fancyreflemlabelprefix}{lem}		%
\newcommand*{\fancyrefcorlabelprefix}{cor}		%
\newcommand*{\fancyrefdeflabelprefix}{def}		%
\newcommand*{\fancyrefproplabelprefix}{prop}	%
\newcommand*{\fancyrefobslabelprefix}{obs}		%
\newcommand*{\fancyrefalglabelprefix}{alg}		%
\newcommand*{\fancyrefasmlabelprefix}{asm}	    %
\newcommand*{\fancyrefasmslabelprefix}{asms}	    %
\newcommand*{\fancyreftbllabelprefix}{tbl}	    %
\newcommand*{\fancyrefestilabelprefix}{esti}	    %

\frefformat{vario}{\fancyrefseclabelprefix}{Section~#1}
\frefformat{vario}{\fancyrefthmlabelprefix}{Theorem~#1}
\frefformat{vario}{\fancyreflemlabelprefix}{Lemma~#1}
\frefformat{vario}{\fancyrefcorlabelprefix}{Corollary~#1}
\frefformat{vario}{\fancyrefdeflabelprefix}{Definition~#1}
\frefformat{vario}{\fancyrefobslabelprefix}{Observation~#1}
\frefformat{vario}{\fancyrefasmlabelprefix}{Assumption~#1}
\frefformat{vario}{\fancyrefasmslabelprefix}{Assumptions~#1}
\frefformat{vario}{\fancyreffiglabelprefix}{Figure~#1}
\frefformat{vario}{\fancyrefapplabelprefix}{Appendix~#1} 
\frefformat{vario}{\fancyrefproplabelprefix}{Proposition~#1}
\frefformat{vario}{\fancyrefalglabelprefix}{Algorithm~#1}
\frefformat{vario}{\fancyrefeqlabelprefix}{(#1)}
\frefformat{vario}{\fancyreftbllabelprefix}{Table~#1}
\frefformat{vario}{\fancyrefestilabelprefix}{Estimator~#1}

%% file: defs.tex
\safemath{\dictab}{[\,\dicta\,\,\dictb\,]}

\safemath{\ysig}{\bmy}
\safemath{\ysighat}{\hat{\ysig}}
\safemath{\ysigdim}{M}
\safemath{\xsig}{\bmx}
\safemath{\xsigdim}{N}
\safemath{\nx}{n_x}
\safemath{\zsig}{\bmz}
\safemath{\zsigdim}{\ysigdim}
\safemath{\rsig}{\bmr}
\safemath{\Adict}{\bA}
\safemath{\Adicttilde}{\widetilde{\Adict}}
\safemath{\Adictdim}{\outputdim\times\xsigdim}
\safemath{\avec}{\bma}
\safemath{\avectilde}{\tilde{\avec}}
\safemath{\Bdict}{\bB}
\safemath{\Bdicttilde}{\widetilde{\Bdict}}
\safemath{\Cdict}{\bC}
\safemath{\cvec}{\bmc}
\safemath{\Ddict}{\bD}
\safemath{\Ddictdim}{\ysigdim\times\xsigdim}
\safemath{\dvec}{\bmd}
\safemath{\Ddicttilde}{\widetilde{\bD}}
\safemath{\Bonb}{\bB}
\safemath{\bvec}{\bmb}
\safemath{\Bonbdim}{\ysigdim\times\ysigdim}
\safemath{\noise}{\bmn}
\safemath{\noisedim}{\ysigim}
\safemath{\err}{\bme}
\safemath{\errdim}{\ysigdim}
\safemath{\errset}{\setE}
\safemath{\nerr}{n_e}
\safemath{\delop}{\bP_\errset}
\safemath{\delopc}{\bP_{{\errset}^c}}

\safemath{\cplxi}{\imath}
\safemath{\cplxj}{\jmath}

\safemath{\dict}{\matD}
\safemath{\inputdim}{N}		%
\safemath{\outputdim}{M}		%
\safemath{\sparsity}{S}	%
\safemath{\inputdimA}{{N_a}}	%
\safemath{\inputdimB}{{N_b}}	%
\safemath{\elemA}{{n_a}}	%
\safemath{\elemB}{{n_b}}	%
\safemath{\resA}{\matR_a}	%
\safemath{\resB}{\matR_b}	%
\safemath{\subD}{\matS} %
\safemath{\subA}{\matS_a} %
\safemath{\subB}{\matS_b} %
\safemath{\dicta}{\matA} 	%
\safemath{\dictb}{\matB} 	%
\safemath{\hollowS}{H}
\safemath{\hollowA}{H_a}
\safemath{\hollowB}{H_b}
\safemath{\cross}{Z}
\safemath{\coh}{\mu_d}			%
\safemath{\coha}{\mu_a}			%
\safemath{\cohb}{\mu_b}			%
\safemath{\mubs}{\nu}	%
\safemath{\cohm}{\mu_m} %
\safemath{\dictset}{\setD}	%
\safemath{\dictsetp}{\dictset(\coh,\coha,\cohb)}	%
\safemath{\dictsetgen}{\dictset_\text{gen}}
\safemath{\dictsetgenp}{\dictsetgen(\coh)}
\safemath{\dictsetonb}{\dictset_\text{onb}}
\safemath{\dictsetonbp}{\dictsetonb(\coh)}

\safemath{\leftside}{U}
\safemath{\rightsideA}{R_a}
\safemath{\rightsideB}{R_b}

\safemath{\indexS}{\setI_S} %

\safemath{\na}{n_a}			%
\safemath{\nb}{n_b}			%
\safemath{\coeffa}{p_i}	%
\safemath{\coeffb}{q_j}	%
\safemath{\seta}{\setP}		%
\safemath{\setb}{\setQ}     %
\safemath{\setw}{\setW}	%
\safemath{\setz}{\setZ}	%
\safemath{\cola}{\veca}		%
\safemath{\colb}{\vecb}		%
\safemath{\cold}{\vecd}		%
\safemath{\inputvec}{\vecx} 	%
\safemath{\error}{\vece}	%
\safemath{\noiseout}{\vecz} 	%
\safemath{\inputvecel}{x}
\safemath{\inputveca}{\vecx_a}
\safemath{\inputvecb}{\vecx_b}
\safemath{\outputvec}{\vecy}	%
\safemath{\lambdamin}{\lambda_{\mathrm{min}}}

\safemath{\elltwo}{\ell_2}
\safemath{\ellone}{\ell_1}
\safemath{\ellzero}{\ell_0}
\safemath{\ellinf}{\ell_\infty}
\safemath{\ellinftilde}{\ell_{\widetilde\infty}}
\safemath{\licard}{Z(\coh,\coha,\cohb)}
\safemath{\xsol}{\hat{x}}
\safemath{\xbord}{x_b}		%
\safemath{\xstat}{x_s}		%
\safemath{\xstatLone}{\tilde{x}_s}
\safemath{\order}{\mathcal{O}} %
\safemath{\scales}{\Theta} %
\safemath{\ones}{\mathbf{1}} %
\safemath{\zeroes}{\mathbf{0}} %
\safemath{\thlone}{\kappa(\coh,\cohb)} %
\safemath{\constoneA}{\delta} %
\safemath{\constoneB}{\epsilon} %
\safemath{\nlarge}{L}				   %
\safemath{\sumlarge}{S_\nlarge}
\safemath{\maxlarger}{P_\nlarge}	   %
\safemath{\Pzero}{\textrm{P0}}	
\safemath{\Pone}{\textrm{P1}}
\safemath{\vecfir}{\vecw}			 %
\safemath{\vecsec}{\vecz}
\safemath{\elvecfir}{w}              %
\safemath{\elvecsec}{z}				 %
\safemath{\nlargefir}{n}
\safemath{\normout}{\gamma}
\safemath{\auxfun}{h}
\safemath{\supp}{\textrm{supp}}%

\safemath{\indexa}{\ell}
\safemath{\indexb}{r}
\safemath{\indexc}{i}
\safemath{\indexd}{j}

\safemath{\project}{P}%

%% file: Introduction.tex
\section{Introduction}
\label{sec:Intro}
Active search \cite{garnett2011bayesian, garnett2012bayesian}, in applications like wildlife patrolling and environment monitoring, involves autonomous robots (agents) discovering objects of interest (OOI) in an unknown environment by adaptively making sequential data collection decisions %
\cite{1291662, 620182, flaspohler2019information}. In such real world scenarios, active search algorithms are faced with two types of uncertainty: %
detection uncertainty and location uncertainty of observed OOIs. %
Due to noise in the sensing setup, an agent may mistake something that is not a target to be an OOI or vice versa, thereby giving rise to detection uncertainty. Further, systematic errors in the sensors may cause an agent to perceive OOIs to be located nearer to or farther from itself than they actually are, leading to uncertainty about the target's true location. Prior research has typically addressed only one of these while assuming no uncertainty in the other.  In robotics, localization and mapping or tracking algorithms consider only target location uncertainty while abstracting away false positive or negative target detections \cite{alsayed2021drone,huang2019survey}. On the other hand, adaptive compressive sensing \cite{4786013,6809991}, Bayesian optimization \cite{rajan2015bayesian,6385653} and bandit style algorithms \cite{abbasi2012online,carpentier2012bandit} only consider uncertainty in target detection while assuming perfect localization. Unfortunately, none of these %
algorithms present a unified framework accounting for both target detection and location uncertainty in active search.

Besides facing observation uncertainty, multi-agent active search introduces an additional challenge. %
Centralized planning with multiple agents is common but often impractical due to communication constraints %
\cite{1291662,yan2013survey}. Further, a system dependant on a central coordinator expecting synchronicity from all agents is also susceptible to communication or agent failure. %
In our problem formulation, the agents share information with their peers when possible, but do not depend on any particular communication arriving. We assume that each agent independently plans and executes its sensing actions using whatever information it already has or receives.%

\fakeparagraph{Contributions}
In this paper, we introduce a realistic sensing model that accounts for uncertainty in both target detection and location. Using these noisy observations, %
we propose an online inference algorithm called UnIK (Uncertainty-aware Inference using Kalman filter) for estimating the presence and position of targets in an unknown environment. %
Using UnIK, we formulate a Thompson sampling (TS) based multi-agent active search algorithm called TS-UnIK to enable decentralized decision making with asynchronous communication. %
We provide simulation results showing that under an identical set of sensing measurements, UnIK significantly outperforms other inference methods that only consider either target detection or location uncertainty.  
We provide simulation results demonstrating the efficiency of the %
action selection strategy in TS-UnIK. %
Finally, we demonstrate TS-UnIK in a realistic simulation environment created on the Unreal (UE4) platform with Airsim plugin. We implement OOI detection and location uncertainty modeling %
and show the real-world transferability of TS-UnIK using a ground robot in this setup. %

%% file: ProblemSetup.tex
\section{Problem Definition}
\label{sec:prob_def}
\begin{figure}[t]%
    \centering
        \begin{subfigure}{0.48\linewidth}%
            \centering
            \includegraphics[scale=0.2]{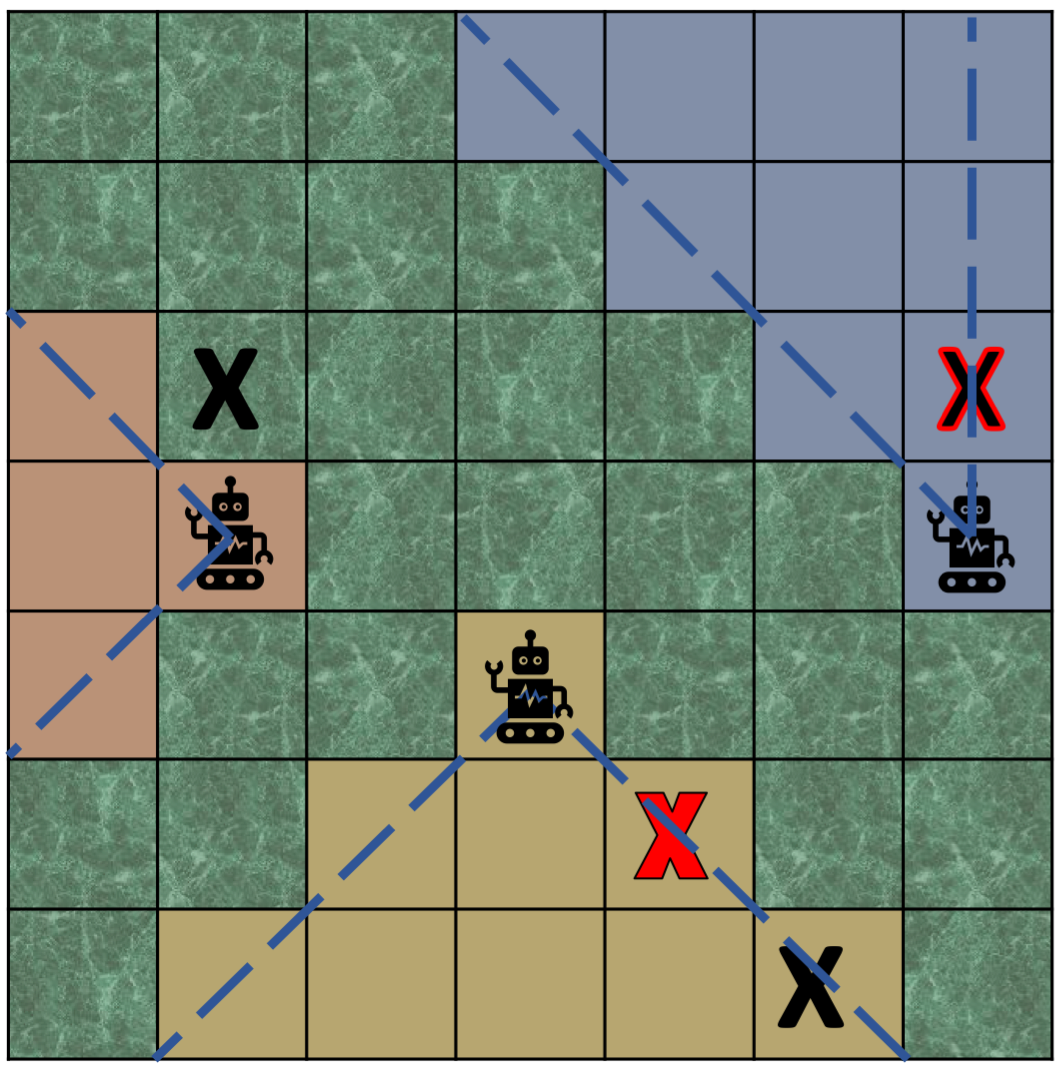}
        	\caption{}
        	\label{fig:1a}
    	\end{subfigure}%
        \begin{subfigure}{0.48\linewidth}
            \centering
            \includegraphics[width=\textwidth]{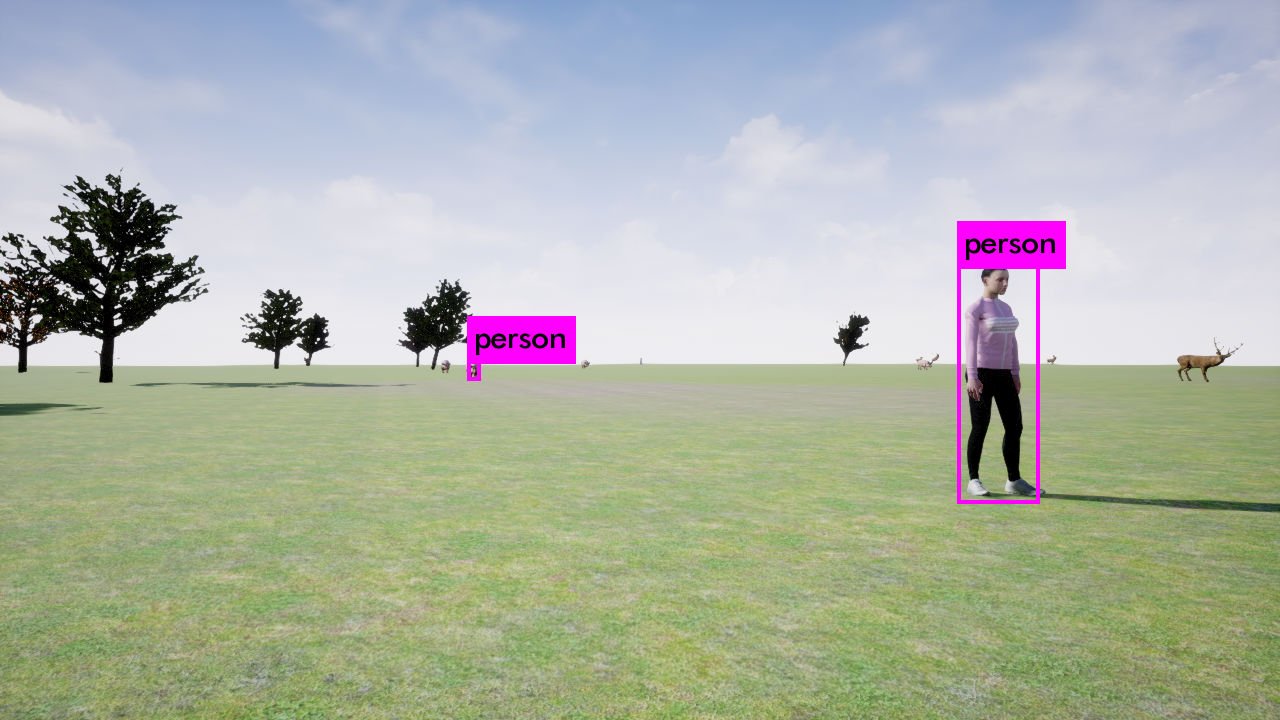}
            \caption{}
            \label{fig:ue4_yolo}
        \end{subfigure}
        \begin{subfigure}{0.48\linewidth}%
            \centering
            \includegraphics[scale=0.35]{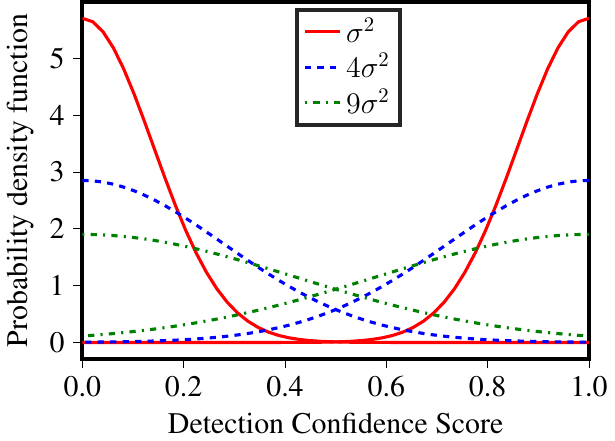}
        	\caption{}
        	\label{fig:det_unc_a}
    	\end{subfigure}%
    	\begin{subfigure}{0.48\linewidth}%
            \centering
            \includegraphics[width=\textwidth]{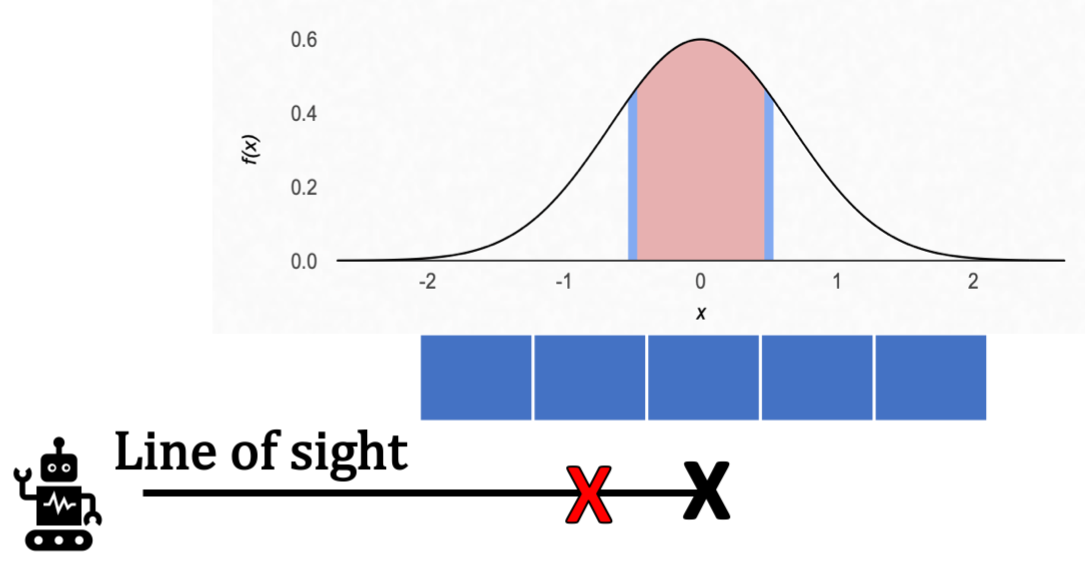}
            \caption{}
            \label{fig:loc_unc_def}
        \end{subfigure}
    \caption{(a) Multiple agents sense different parts of the environment looking for OOIs. True OOIs are crossed in black. Targets detected by the agent in its field of view are crossed in red. (b) OOI detections using YOLOv3 \cite{redmon2018yolov3} in the robot's FOV in our UE4 simulator (\cref{subsec:UE4}). (c) An object detector becomes less confident about positive (1) as well as negative (0) labels as its distance to the object increases (\cref{eq:detection_uncertainty}).
    (d) Due to location uncertainty, the observed depth of the true OOI (``X" in black) is shifted towards the agent, to the grid cell marked ``X" in red. %
    }
    \label{fig:1}
\end{figure}

Consider multiple ground robots searching a space to locate some OOIs %
(\cref{fig:1a}). %
Each robot moves around and observes certain regions of the search space by taking pictures to detect OOIs 
and measuring their distances from its current location. 
We assume the robots can localize themselves accurately.  The colored cells in \cref{fig:1a} illustrate each robot's sensing action using a camera with a 90$\degree$ field of view (FOV). Each robot independently decides its next sensing action given its current belief about OOIs in the search space. 

Once a robot senses a region, it obtains information about both the presence (detection) and corresponding depth (location) of objects in its FOV. For example, RGB-D sensors coupled with an object detector like YOLOv3 \cite{redmon2018yolov3} can be used to extract such information. The object detector identifies OOIs with a confidence score that varies with distance from the camera. Objects farther away from the camera usually have a lower probability of being correctly identified. This gives rise to target detection uncertainty in the robot's observation. Further, the depth %
sensor is prone to error in the measured distance to the OOI.  This leads to target location uncertainty in the robot's observation. Our objective is to account for both the detection and location uncertainty in our observation model and utilize it to make improved active sensing decisions.%

\subsection{Ground truth model for target detection uncertainty}
\label{subsec:det_unc}
Let $\xi_i \in \{0,1\}$ denote the output of an object detector with perfect detection ability which labels an object $i$ at a distance $l_i$ away from the camera accurately with either a `0' (not OOI) or a `1' (OOI). Typically, the confidence score of the object detector gradually declines as a function of the OOI's distance from the camera. Prior work \cite{ghods2021multi} has characterized this behavior using a depth aware detection model where 
the output $y_i$ %
of an imperfect object detector is the true object label modified by an additive one-sided Gaussian noise:%
\begin{align}\label{eq:detection_uncertainty}
y_i = \xi_i + n^d_i,\;\text{with}\;n^d_i \sim \normal^+(0,\sigma_i^2(l_i)).
\end{align} 
Therefore, $y_i \in [0,1]$ follows a one-sided normal distribution centered at $\xi_i$$=$$0$ for a true negative and $\xi_i$$=$$1$ for a true positive label. 
The variance $\sigma_i^2(\cdot)$ is an increasing function of $l_i$.
So the probability of a false negative or a false positive OOI detection increases at distances farther away from the camera. This is illustrated in \cref{fig:det_unc_a} where the horizontal axis indicates the detector's $y_i$ and the %
curves indicate the varying probability densities at different object distances $l_i$.

\subsection{Ground truth model for target location uncertainty}
\label{subsec:loc_unc}
Let $\zeta_i\in\mathbb{R}$ denote the measurement from an accurate depth sensor of its true distance to an object $i$. Real depth measurements have error along the agent's line of sight. Prior work \cite{belhedi2012noise} has
characterized the error %
as a Gaussian distribution. %
In our setup, we model the target location uncertainty as additive Gaussian noise %
so that %
\begin{align}\label{eq:location_uncertainty}
    y_i = \zeta_i + n^\ell_i,\;\text{with}\;n^\ell_i \sim \normal(0, r_u^2)
\end{align}
where $y_i\in\mathbb{R}$ is the measurement from an imperfect depth sensor and 
$r_u$ parameterizes the uncertainty of such measurements along the agent's line of sight. As depicted in \cref{fig:loc_unc_def}, the measured depth follows a normal probability distribution centered on the true target location. %
\subsection{Problem Formulation}
\label{subsec:prob_form}
We consider a gridded search environment described by a sparse matrix which we want to recover through multi-agent active search. $M$ is the total number of grid cells. $\bm\beta\in\{0,1\}^{M\times 1}$ denotes the flattened vector representation of the environment having $k$ non-zero entries at the true locations of the $k$ OOIs. $\bX_t \in \{0,1\}^{Q_t\times M}$ is the sensing action at time $t$. $Q_t$ is the number of grid cells covered by the robot's FOV under $\bX_t$. %
Each row of $\bX_t$ is a one hot vector $\{0,1\}^{1\times M}$ indicating the position of one of the sensed grid cells in the robot's FOV. $\bX_t\bm\beta\in\{0,1\}^{Q_t\times1}$ is the ground truth observation due to sensing action $\bX_t$. $\bmy_t\in\mathbb{R}^{Q_t\times 1}$ is the agent's observation vector indicating the noisy sensor measurement %
from executing $\bX_t$. The sensing model is %
\begin{align}\label{eq:sensing_model}
    \underbrace{\bmy_t}_{\text{noisy measurement}} = \underbrace{\bX_t\bm\beta}_{\text{ground truth observation}} + \underbrace{\bmn_t}_{\text{measurement noise}}
\end{align}
where $\bmn_t \in \mathbb{R}^{Q_t\times 1}$ is composed of the noise from (\ref{eq:detection_uncertainty}), %
(\ref{eq:location_uncertainty}).

\fakeparagraph{Remark 1.} Note that the model described above is what our simulator considers to be ground truth. %
Our algorithm and its agents are neither aware of the number of targets nor their true locations, and only receive the measurement $(\bX_t,\bmy_t)$.%

\fakeparagraph{Communication:} We assume that communication, although unreliable, will be available sometimes and the agents should %
communicate when possible. The agents share their measurements asynchronously, and do not wait on inter-agent communications. %
Further, the set of available past measurements need not remain consistent across agents due to communication unreliability. %

\fakeparagraph{Remark 2.} Since our goal is %
uncertainty-aware active sensing and not %
planning a continuous path, we only require a coarse discretization of our environment. For example, in \cref{subsec:UE4}, we cover a 250m$\times$250m region with square grid cells of size 15m. We also assume that individual OOIs occupy an entire grid cell and the location uncertainty from the depth sensor only affects which cell was determined to have the OOI, not the OOI's placement within the cell.

To recover the search vector $\bm\beta$ by actively identifying all the OOIs, at each time $t$, an agent $j$ chooses the best sensing action $\bX_t^j$ based on its belief about the OOIs given the measurements available thus far in its measurement set $\bD_t^j$. Assuming that all the agents collectively obtain $T$ measurements, our objective is to correctly estimate the sparse vector $\bm\beta$ with as few measurements $T$ as possible.
For a single agent, our problem reduces to sequential decision making with the measurement set $\bD_t^1 = \{(\bX_1,\bmy_1),\dots,(\bX_{t-1},\bmy_{t-1})\}$ available to the agent at time $t$. In the multi-agent setting, following our communication constraints, we use a decentralized and asynchronous parallel approach with %
agents independently deciding individual sensing actions \cite{kandasamy2018parallelised, ghods2021multi, ghods2021decentralized}.

%% file: RelatedWorks.tex
\section{Related Work}
\label{sec:related_work}

The issue of location uncertainty has been studied by the robotics community, particularly in tasks such as path planning, localization and tracking \cite{1633413,932739}.
Such problems are typically approached using filtering algorithms that can recursively estimate the robot or target state and the associated location uncertainty, while abstracting away the detection uncertainty by thresholding the detection probability to zero or one. 
Sensor measurements from RGB cameras and depth sensors used in typical robotic tasks like SLAM and navigation often introduce data association error due to pose mismatch between the corresponding sensors \cite{basso2018robust}.
Further, depth images suffer from position dependent geometric distortions and distance dependent measurement bias \cite{belhedi2012noise}. This also introduces location uncertainty in the sensor measurements. As a result, %
sensor calibration algorithms have been proposed which %
parameterize the error in the depth measurements and then learn %
those parameters from carefully collected training data %
\cite{zhou2014simultaneous, zuniga2019intrinsic, 8588983, 6696737, 6907780}. 
But they typically engineer away the need to account for detection uncertainty by using some predetermined visual patterns in the collected dataset.
Unlike our problem setup, they do not focus on learning how to autonomously collect appropriate data %
to reduce location uncertainty in the sensor measurements.  

The domain of sparse signal recovery, on the other hand, is primarily concerned with the problem of choosing sensing actions in the face of detection uncertainty, assuming that once the signal is detected then its location is accurate. Prior work in this area has proposed algorithms that use principles from compressed sensing together with constrained optimization to estimate the signal \cite{5356153,5419092}. \cite{ma2017active} formulated this as an active search problem with realistic region sensing actions %
and \cite{9632368} proposed a reinforcement learning approach incorporating the detection uncertainty in the observations into a POMDP framework.
\cite{ghods2021decentralized,ghods2021multi} build on the former setup and go on to model detection uncertainty as a function of the target's distance from the sensing agent. 

The discrete spatial search problem is also studied in search theory by considering detection uncertainty through false positive detections in sensing individual cells \cite{kress2008optimal,6059507,cheng2019scheduling} but they do not relate the uncertainty with sensor capabilities and do not generalize to realistic region sensing actions. 

In terms of the agent's decision making, active search has close similarities with the task of information gathering for planning and navigation in robotics. The majority of such algorithms use entropy and information gain as an optimization objective for deciding which action to execute \cite{rajan2015bayesian,ma2017active}. Unfortunately this leads to deterministic action selection for an information greedy agent. As a result, much of the work in this domain has generally focused on single agent settings \cite{cliff2015online,lim2016adaptive,patten2018monte}, and if multi-agent, they typically rely on the presence of a centralized controller to coordinate the agents so that they can avoid all choosing the same action at any decision making step \cite{arora2019multi,surmann2019integration}. Recently, motivated by the real world constraints of centralized multi-agent systems, %
some algorithms have been proposed with a centralized planning followed by decentralized execution approach \cite{gupta2017cooperative,dames2017detecting,iqbal2019actor}. Further, some others propose to achieve decentralization by having the agents repeatedly communicate their future plans (e.g. beliefs or intended sensing actions) with their peers \cite{best2019dec,best2020decentralised}. This can become cumbersome especially if the belief is high dimensional and there is unreliable inter-agent communication. In contrast, \cite{ghods2021decentralized,ghods2021multi} consider a realistic setup and introduce posterior sampling based active search in a decentralized multi-agent system with unreliable at-will communication of sensing measurements among peers. %

The task of characterizing the uncertainty of detection and location of objects in an image has also been extensively studied in computer vision \cite{Hall_2020_WACV,7789580,gonzalez2015active,caicedo2015active,9340798}. In contrast, our %
agents adaptively \emph{choose} which images to capture (i.e. decide where to sense and in which direction) by considering the associated uncertainties %
to recover all OOIs in a search space.

%% file: Algorithm.tex
\section{Our proposed algorithms: UnIK and TS-UnIK}
\label{sec:algo}
We now describe our approach to the multi-agent active search problem described in \cref{subsec:prob_form} in two stages. First, we outline our inference procedure that agents use to identify OOIs and estimate their locations given the set of available measurements thus far. Next, we describe our decentralized and asynchronous multi-agent decision making algorithm that %
utilizes the estimates from the proposed inference method.  
\subsection{Proposed inference algorithm: UnIK}
\label{subsec:UNIK}
Following \cref{subsec:prob_form}, we want to recover $\bm\beta$ by identifying all the OOIs. But both the number and position of targets are unknown to the agents. 
So we initialize each agent with a Gaussian prior over $\bm\beta$, denoted by $p_0(\bm\beta) = \normal{(\hat{\bm\beta_0},\bP_0)}$. Given the measurement set $\bD_t^j$ available to the agent $j$ at time $t$, it updates its posterior belief $p(\bm\beta|\bD_t^j) = \normal(\hat{\bm\beta}_t^j,{\bP}_t^j)$ using a Kalman filter (KF) \cite{kalman1960new} with process model $\bm\beta_t = \bm\beta_{t-1}$ and measurement model $\bmz_t = \bX_t\bm\beta_t + \bm\nu_t$, where $\bm\nu_t\sim\normal(\mathbf{0},\mathbf{\Sigma}_{\bmz_t})$. 
The prediction and update steps of the KF are as follows.%
\begin{align}
    &\text{\emph{Prediction}: }\hat{\bm\beta}_t^- = \hat{\bm\beta}_{t-1},\;\bP_t^- = \bP_{t-1}\\
    &\text{\emph{Kalman gain}: }\bK_t = \bP_t^-\bX_t^T(\bX_t\bP_t^-\bX_t^T + \mathbf{\Sigma}_{\bmz_t} + \lambda_t\mathbf{I})^{-1}\\
    &\text{\emph{Update}: }\hat{\bm\beta}_t = {\hat{\bm\beta}}_t^- + \bK_t(\bmy_t - \bX_t{\hat{\bm\beta}}_t^-)\label{eq:state_update}\\
    &\bP_t = (\mathbf{I} - \bK_t\bX_t)\bP_t^-(\mathbf{I} - \bK_t\bX_t)^T + \bK_t\mathbf{\Sigma}_{\bmz_t}\bK_t^T
\end{align}

$\lambda_t$ is a regularization constant. Given sensing action $\bX_t$ and observation $\bmy_t$ (\ref{eq:sensing_model}), the agent constructs measurement noise covariance matrix $\mathbf{\Sigma}_{\bmz_t}$ to incorporate both target detection and location uncertainty in the KF measurement model.
First, $\mathbf{\Sigma}_{\bmz_t} = \text{diag}([\sigma_q^2(l_q)]_{q=1}^{Q_t})$ is initialized to account for the OOI detection uncertainty in the robot's FOV using the distance dependent variance (\cref{subsec:det_unc}).  Next, the agent estimates the possible OOI locations in its FOV by thresholding %
$\bmy_t$: $\bmy_{thr,t} = \mathbf{1}(\bmy_t >= c_{thr})$. For each possible 
OOI in $\bmy_{thr,t}$, the agent then determines its location uncertainty field i.e. the grid cells in its FOV where this OOI might be truly located. 

\begin{figure}[htp]
    \centering
    \begin{subfigure}{0.5\linewidth}
    \includegraphics[scale=0.25]{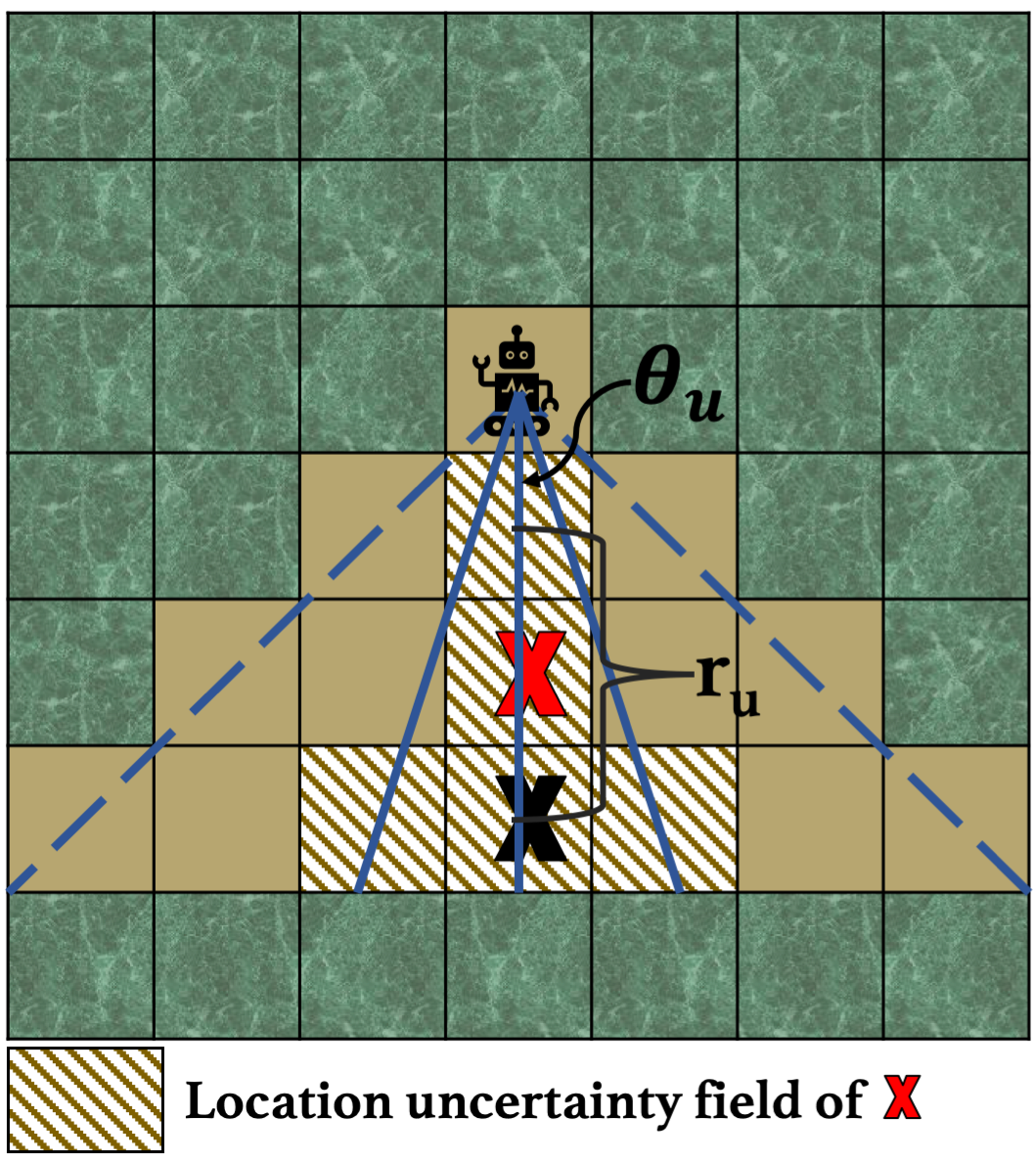}
    \caption{}
    \label{fig:locunc_model}
    \end{subfigure}%
    \begin{subfigure}{0.5\linewidth}
    \includegraphics[scale=0.5]{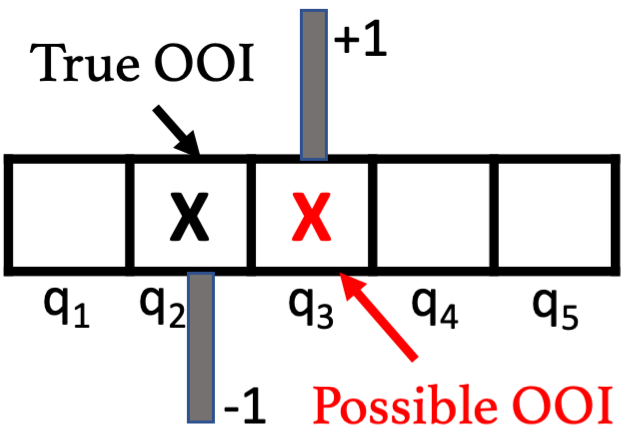}
    \caption{}
    \label{fig:locunc_pmf}
    \end{subfigure}
    \caption{(a) Showing the location uncertainty field for an OOI detected in $\bmy_{thr,t}$ at the position marked ``X" in red. (b) Showing part of the additive noise vector due to target location uncertainty in an agent's observation (\ref{eq:sensing_model}) when an OOI is actually present at $q_2$ but observed at $q_3$.}
\end{figure}

Following \cref{fig:locunc_model}, suppose $\bmy_{thr,t}$ indicates an OOI in the position marked ``X" in red, whereas the OOI is actually located in the grid cell marked ``X" in black. Along its line of sight to the observed OOI location (``X" in red), the agent computes the location uncertainty field using the detector's radial uncertainty parameter $r_u$ (same as (\ref{eq:location_uncertainty})) and an angular uncertainty parameter $\theta_u$ shown in \cref{fig:locunc_model}. It will include all the grid cells in the region bounded radially by [$-r_u,r_u$] around the observed OOI location within an angular spread of [$-\theta_u,\theta_u$] relative to the line of sight to the agent. 
Next, the location uncertainty field for each possible OOI index in $\bmy_{thr,t}$ is used to compute the variance and covariance due to location uncertainty in $\mathbf{\Sigma}_{\bmz_t}$. %
\begin{table}[htp]
    \centering
    \caption{Modeling the probability distribution for target location uncertainty in the measurement model using the observed target location at $q_3$ and its uncertainty field $\{q_1,q_2,q_3,q_4,q_5\}$.}
    \resizebox{0.85\columnwidth}{!}{%
    \begin{tabular}{c|c|lcccccll}
    \toprule
    $\tilde{q}_{OOI}$ & $q_{OOI}$ & \multicolumn{8}{c}{$\bmn_{t,q_{OOI}}^\ell$} \\
    \midrule
    $q_3$ & $q_1$ & [ & -1 & 0 & 1 & 0 & 0 &$\cdots$ & $0]^T$\\ 
     & $q_2$ & [ & 0 & -1 & 1 & 0 & 0 &$\cdots$ & $0]^T$\\
     & $q_3$ & [ & 0 & 0 & 0 & 0 & 0 &$\cdots$ & $0]^T$\\
     & $q_4$ & [ & 0 & 0 & 1 & -1 & 0 &$\cdots$ & $0]^T$\\
     & $q_5$ & [ & 0 & 0 & 1 & 0 & -1 &$\cdots$ & $0]^T$\\
    \bottomrule
    \end{tabular}
    }
    \label{table:probability_model}
\end{table}

Using the computed location uncertainty field for an OOI at $\tilde{q}_{OOI}$ in its FOV, the agent computes a probability function over the grid cells where the OOI could truly be located. Referring to the observation noise due to location uncertainty $\bmn_t^\ell$ in \cref{subsec:loc_unc}, we note that in a discretized grid environment, it essentially results in swapping the OOI's observed location within a region around the true OOI location. This needs to be captured in the measurement model for the agent's KF. For example, following \cref{fig:locunc_pmf}, if $\bmy_{thr,t}$ indicates that an OOI is located at $\tilde{q}_{OOI}=q_3$ and its location uncertainty field is $\{q_1,q_2,q_3,q_4,q_5\}$, \cref{table:probability_model} shows the possible location swaps the agent reasons about including the case where the OOI's true location is also $q_3$. 
In \cref{table:probability_model}, $\tilde{q}_{OOI}$ is the observed location, $q_{OOI}$ is the possible true location and $\bmn_{t,q_{OOI}}^l$ is the noise component due to location uncertainty that executes the position swap between the true and observed OOI locations. The index with $-1$ indicates $q_{OOI}$ and that with $1$ indicates $\tilde{q}_{OOI}$.
After enumerating all feasible OOI position swaps in the location uncertainty field, %
the agent updates the measurement noise covariance as $\mathbf{\Sigma}_{\bmz_t}^{q_3,q_3} \pluseq 0^2\times\frac{1}{5}+(1)^2\times(1-\frac{1}{5})$ and $\mathbf{\Sigma}_{\bmz_t}^{q_{1/2/4/5},q_{3}} = \mathbf{\Sigma}_{\bmz_t}^{q_3,q_{1/2/4/5}} \pluseq 1\times(-1)\times\frac{1}{5}+1\times0\times\frac{3}{5}+0\times0\times\frac{1}{5}$. This is repeated for all possible OOIs in the agent's FOV. The constructed measurement noise covariance $\mathbf{\Sigma}_{\bmz_t}$ is then used to update the agent's KF with $(\bX_t, \bmy_t)$. 

\begin{algorithm}[htp]
	\caption{UnIK (inference with random action selection)}
		\begin{algorithmic}[1]
            \State{\textbf{Assume:}Sensing model (\ref{eq:sensing_model}), sparse true state $\bm\beta$}
            \State{$\bD_0^j=\phi$, $p_0^j(\bm\beta) = \normal(\hat{\bm\beta}_0^j,\bP_0^j)$, $\hat{\bm\beta}_0^j = \frac{\mathbf{1}_{M\times1}}{M}$, $\bP_0^j = \sigma_0^2\mathbf{I}$}
			\For{$t$ in $\{1,\hdots,T\}$}
            \State{Select $\bX_t^j$ uniform randomly; observe $\bmy_t^j$.}
			\State{Update $\bD_t^j = \bD_{t-1}^j\cup\{(\bX_t^j,\bmy_t^j)\}$}\label{line:6}
			\State{Obtain $\bmy_{thr,t}^j = \mathbf{1}(\bmy_t^j\geq c_{thr})$}\label{line:UnIK_thresh}
			\State{Initialize $\mathbf{\Sigma}_{\bmz_t}$ to account for detection uncertainty}%
            \For{each possible OOI in $\bmy_{thr,t}^j$}\Comment{(\cref{subsec:UNIK})}\label{line:UnIK_sigma} %
			\State{Obtain its location uncertainty field; update $\mathbf{\Sigma}_{\bmz_t}$}
			\EndFor
			\State{Estimate $\hat{\bm\beta}_t^j, \bP_t^j$ using KF}\label{line:UnIK_estimate}
			\EndFor
	    \end{algorithmic}
\label{algo:UnIK}
\end{algorithm}

\subsection{Proposed action selection algorithm: TS-UnIK}
\label{subsec:TS-UNIK}
In order to develop an action selection strategy for our agents, we observe that our active search formulation falls within the problem setup for the Myopic Posterior Sampling (MPS) framework developed in \cite{kandasamy2019myopic}. Like MPS, our objective is to actively learn $\bm\beta$ %
with as few measurements as possible. MPS chooses the action that maximizes the one-step lookahead (myopic) reward assuming that a sample drawn from the posterior is the true state of the world. This idea is a generalization of Thompson sampling (TS) \cite{thompson1933likelihood} which is commonly used in Bayesian optimization. Furthermore, parallelized asynchronous TS is an excellent candidate for a decentralized multi-agent algorithm \cite{kandasamy2018parallelised,ghods2021multi,ghods2021decentralized}. By using a posterior sample in the reward function, TS (as well as MPS) enables multiple agents to independently choose distinct sensing actions that maximize their respective rewards while incurring little regret compared to a centralized planner. Therefore we adopt a TS approach in our proposed algorithm. 

Next we describe our TS based reward formulation. At time $t$, an agent $j$ having available measurements $\bD_{t-1}^j$ %
estimates its posterior belief $p(\bm\beta|\bD_t^j) = \normal(\hat{\bm\beta}_{t-1}^j,\bP_{t-1}^j)$ using UnIK. It then draws a sample $\tilde{\bm\beta}_t^j \sim p(\bm\beta|\bD_t^j)$. %
Assuming $\tilde{\bm\beta}_t^j$ to be the true search vector, the agent will prefer choosing an action $\bX_t^j$ that will allow its updated estimate $\hat{\bm\beta}_t^j$ at time $t$ to be as close as possible to $\tilde{\bm\beta}_t^j$. In particular, we want to maximize %
$$\mathcal{R}(\tilde{\bm\beta}_t^j,\bD_{t-1}^j,\bX_t^j) = \frac{\mathbb{E}_{\bmz_t^j|\tilde{\bm\beta}_t^j,\bX_t^j}[-||\tilde{\bm\beta}_t^j - \hat{\bm\beta}_t^j||_2^2]}{\mathbb{E}_{\bmz_t^j|\tilde{\bm\beta}_t^j,\bX_t^j}[||\hat{\bm\beta}_t^j||_2^2]}$$ 
where $\bmz_t^j$ is the KF measurement variable. 
Note that $\hat{\bm\beta}_t^j$ is a one-step lookahead estimate assuming that $\bX_t^j$ would result in an observation $\bmz_t^j$ following the KF measurement model if $\bm\beta_t^j = \tilde{\bm\beta}_t^j$. 
The numerator favours actions $\bX_t^j$ such that in expectation, the estimate $\hat{\bm\beta}_t^j$ is close to the assumed true state $\tilde{\bm\beta}_t^j$. The denominator $\mathbb{E}_{\bmz_t^j|\tilde{\bm\beta}_t^j,\bX_t^j}[||\hat{\bm\beta}_t^j||_2^2]$ is analogous to the estimated strength of the signal, and since the numerator is non-positive, a larger denominator will lead to an overall higher reward. Therefore, our formulated reward prefers sensing actions that would maximally reduce the uncertainty in the agent's current posterior distribution. %

Using $\hat{\bm\beta}_t^j = \hat{\bm\beta}_{t-1}^j + \bK_t^j(\bmz_t^j - \bX_t^j\hat{\bm\beta}_{t-1}^j)$ and $\bmz_t^j \sim \normal(\bX_t^j\tilde{\bm\beta}_t^j,\bm\Sigma_{\bmz_t^j})$, we can compute $\mathcal{R}(\tilde{\bm\beta}_t^j,\bD_{t-1}^j,\bX_t^j)$ as:
\begin{align}
    \frac{\splitfrac{
        -||\tilde{\bm\beta}_t^j - \hat{\bm\beta}_{t-1}^j + \bK_t^j\bX_t^j\hat{\bm\beta}_{t-1}^j||_2^2 
    + 2(\tilde{\bm\beta}_t^j - \hat{\bm\beta}_{t-1}^j}{+ \bK_t^j\bX_t^j\hat{\bm\beta}_{t-1}^j)^T\bK_t^j\bX_t^j\tilde{\bm\beta}_t^j 
    - ||\bK_t^j||_2^2(\text{tr}(\bm\Sigma_{\bmz_t^j}) + ||\bX_t^j\tilde{\bm\beta}_t^j||_2^2)}}{\splitfrac{||\hat{\bm\beta}_{t-1}^j - \bK_t^j\bX_t^j\hat{\bm\beta}_{t-1}^j||_2^2 
    + ||\bK_t^j||_2^2(\text{tr}(\bm\Sigma_{\bmz_t^j}) 
    + ||\bX_t^j\tilde{\bm\beta}_t^j||_2^2)}{
    + 2(\hat{\bm\beta}_{t-1}^j - \bK_t^j\bX_t^j\tilde{\bm\beta}_t^j)^T\bK_t^j\bX_t^j\tilde{\bm\beta}_t^j
    }}\label{eq:reward}
\end{align}

Finally, among all actions $\{\bX_t^j\}$ at time $t$, agent $j$ chooses ${\bX_t^j}^*|\tilde{\bm\beta}_t^j = \argmax_{\bX_t^j} \mathcal{R}(\tilde{\bm\beta}_t^j,\bD_{t-1}^j,\bX_t^j)$. %

\begin{algorithm}[htp]
	\caption{TS-UnIK}
		\begin{algorithmic}[1]
            \State{\textbf{Assume:}Sensing model (\ref{eq:sensing_model}), sparse true state $\bm\beta$, $J$ agents}
            \State{$\bD_0^j$$=$$\phi$, $p_0^j(\bm\beta)$$=$$\normal(\hat{\bm\beta}_0^j,\bP_0^j)$, $\hat{\bm\beta}_0^j$$=$$\frac{1}{M}\mathbf{1}_{M\times1}$, $\bP_0^j$$=$$\sigma_0^2\mathbf{I}$}%
			\For{$t$ in $\{1,\hdots,T\}$}\Comment{For any available agent $j$}
			\State{Sample $\tilde{\bm\beta}_t^j \sim p(\bm\beta | \bD_{t-1}^j) = \normal(\hat{\bm\beta}_{t-1}^j, \bP_{t-1}^j)$}
			\State{${\bX_t^j}^* = \argmax_{\bX_t^j}\mathcal{R}(\tilde{\bm\beta}_t^j,\bD_{t-1}^j,\bX_t^j)$. Observe $\bmy_t^j$.}\label{line:5_TS}%
			\State{Update $\bD_t^j = \bD_{t-1}^j\cup\{({\bX_t^j}^*,\bmy_t^j)\}$. Share $({\bX_t^j}^*,\bmy_t^j)$ asynchronously with teammates.}
			\State{Estimate $\hat{\bm\beta}_t^j, \bP_t^j$ using UnIK ( \cref{line:UnIK_thresh}-\cref{line:UnIK_estimate})} %
			\EndFor
	    \end{algorithmic}
\label{algo:app_TS-UnIK}
\end{algorithm}

\fakeparagraph{Remark 3.} TS based decision making enables each agent to independently choose its next sensing action based on the uncertainty in its current posterior over the search space, and subsequently update its individual posterior estimates using its own measurements and those from other agents. We do not require perfect data association between measurements from different agents on the same OOI, nor a central controller for synchronization of observations across agents. %

\fakeparagraph{Computational complexity}: Due to the Kalman filter based recursive nature of UnIK, for any agent, the time complexity of each inference step is the same and bounded by O($M^{2.376}$). %
Additionally, TS-UnIK requires an agent to select the best (maximum reward) among all feasible actions at each time step. Therefore \cref{line:5_TS} in \cref{algo:app_TS-UnIK} results in O($|\mathcal{A}|M^2$) complexity %
where $|\mathcal{A}|$ is the size of the agent's %
action space. 

%% file: Results.tex
\section{Experimental Results}
\label{sec:results}
We now describe the simulation setup for our experimental results. %
Consider a 2-dimensional (2D) discretized search space having $J$ agents tasked with recovering the unknown search vector $\bm\beta$ which is randomly generated using a uniform sparse prior with $k$ non-zero entries each with value 1. Note that the agents are not aware of the number of targets (OOIs) $k$. We assume that agents are positioned at the centre of the grid cells they occupy and are free to move in any direction in the search space. When positioned in a cell, %
an agent can look in one of 4 possible directions: north, south, east or west to a maximum distance of 5 grid cells ahead with a $90\degree$ FOV. 

In the following experiments, the agents' performance is measured using the full recovery rate which is defined as the rate at which the agents correctly recover the \emph{entire} vector $\bm\beta$. The plots show mean values with shaded regions indicating standard error over multiple trials, each trial differing only in the instantiation of the true position of the $k$ OOIs in $\bm\beta$.

\subsection{Comparing inference}\label{subsec:results_inf}
\begin{figure}[htp]
  \centering
    \begin{subfigure}{0.5\linewidth}
    \centering
    \includegraphics[width=\textwidth]{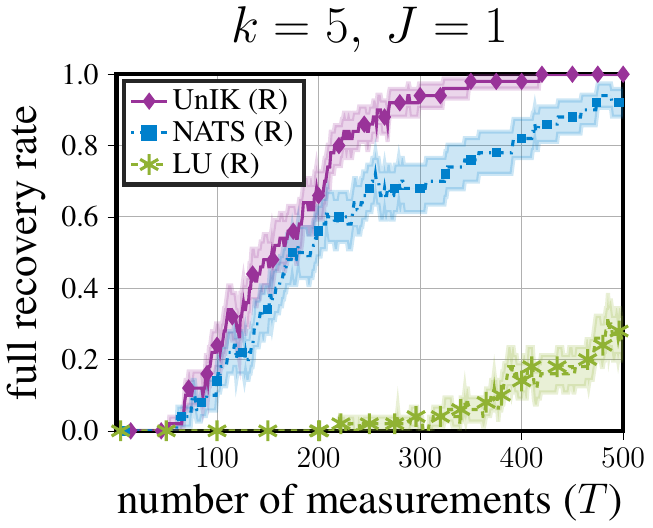}
    \caption{}
    \label{fig:inf_p1}
    \end{subfigure}%
    \begin{subfigure}{0.5\linewidth}
    \centering
    \includegraphics[width=\textwidth]{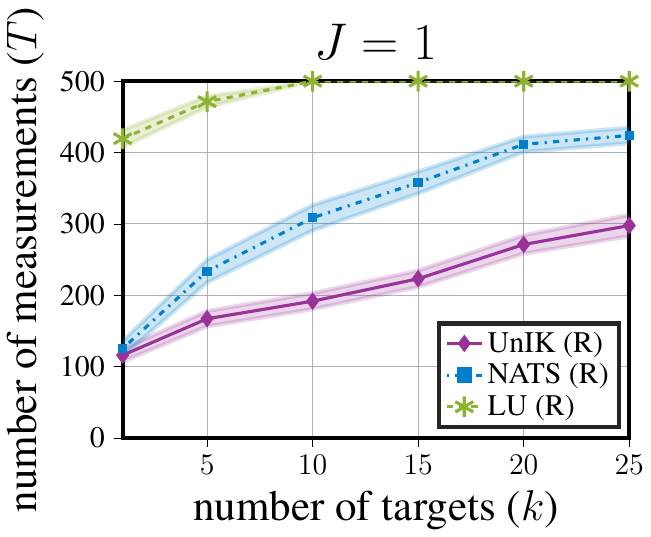}
    \caption{}
    \label{fig:inf_p2}
    \end{subfigure}
    \caption{\textbf{$J=1$ agent in a $16\times16$ grid, $k$ targets}. Mean and standard error over 50 trials. (R) indicates uniformly random action selection. UnIK outperforms baseline inference algorithms that account for only one of detection and location uncertainty, for different number of targets.}
    \label{fig:results_inf}
\end{figure}

We compare UnIK against two other inference methods: (1) NATS \cite{ghods2021multi}, which only accounts for detection uncertainty and (2) LU, which is our designed location uncertainty only baseline. 
\cref{fig:results_inf} illustrates the performance of UnIK compared to NATS and LU in a single agent setting ($J = 1$) in a $16\times16$ search space, over 50 random trials. %
At each step, an action is chosen uniformly at random. %
The agent's actions and corresponding observations are the same across the three inference algorithms. 

\cref{fig:inf_p1} shows the full recovery rate of an agent when there are $k$$=$$5$ targets %
in a $16$$\times$$ 16$ search space. %
Unlike NATS or LU, UnIK leverages the combined target detection and location uncertainty to detect all the OOIs with fewer measurements. %
\cref{fig:inf_p2} further compares the efficiency of the algorithms in terms of number of measurements needed to achieve full recovery by varying the number of targets $k \in \{1,5,10,15,20,25\}$. With increasing $k$, NATS lacking the target location uncertainty modeling requires an increasingly larger number of measurements compared to UnIK. On the other hand, in the absence of target detection uncertainty modeling, LU's performance is sensitive to the detection threshold and declines faster, failing in all 50 trials to recover all OOIs within $T = 500$ measurements for $k > 5$. %

\subsection{Comparing action selection}\label{subsec:results_actsel}
We now demonstrate the efficiency of TS based action selection in combination with our proposed inference method for decentralized and asynchronous multi-agent active search. 

\fakeparagraph{UnIK (R) vs. TS-UnIK}: Following our observations in \cref{subsec:results_inf} where UnIK outperformed other inference algorithms in fully recovering all OOIs, we will now use UnIK with uniformly random action selection (i.e. UnIK (R)) as the baseline against which we compare the performance of TS-UnIK. 
\begin{figure}[htp]
    \centering
    \begin{subfigure}{0.5\linewidth}
        \includegraphics[width=\textwidth]{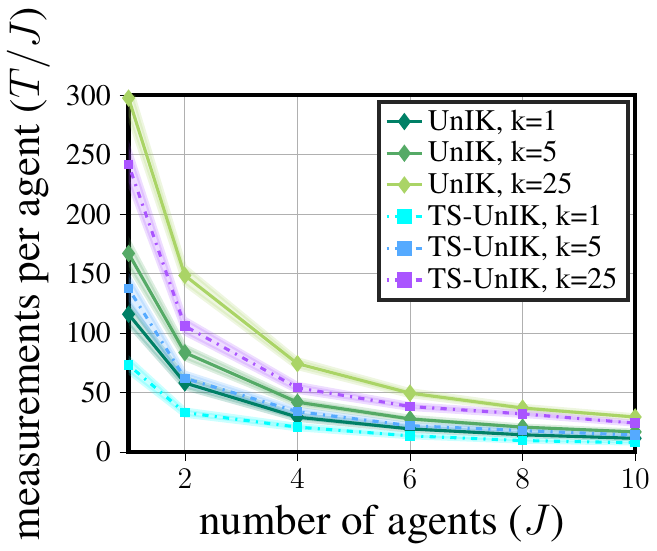}
        \caption{}
        \label{fig:act_vs_k}
    \end{subfigure}%
    \begin{subfigure}{0.5\linewidth}
        \includegraphics[width=\textwidth]{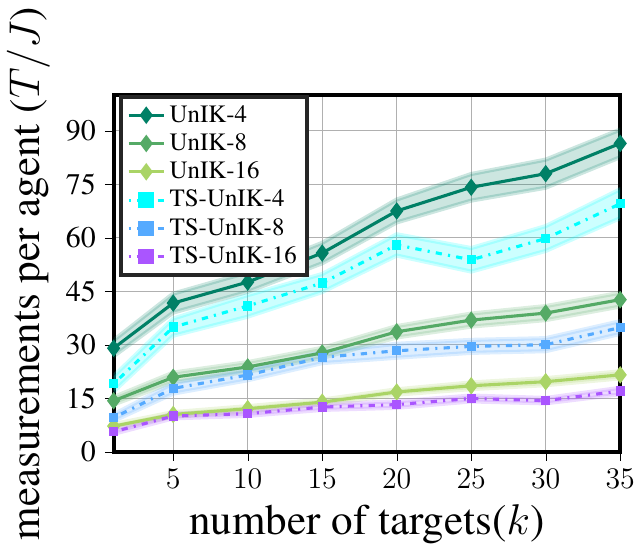}
        \caption{}
        \label{fig:act_vs_g}
    \end{subfigure}
    \caption{TS-UnIK scales better with number of agents $J$ and targets $k$ compared to UnIK (R) in a $16\times16$ grid, 50 trials.}
    \label{fig:results_actsel_vs_g_vs_k}
\end{figure}
\cref{fig:act_vs_k} plots the number of measurements needed per agent ($T/J$) to fully recover $k$ OOIs as the number of agents $J$ increases. In a perfect sensing setup, we would expect a single agent algorithm using the TS based action selection strategy to require $J$ times as many measurements as %
that required per agent in a team with $J$ agents \cite{ghods2021decentralized}. We observe this to hold for TS-UnIK, %
resulting in a $J$ times improvement in performance when the number of agents multiplies $J$ times. As the team size becomes larger, we observe that the full recovery performance plateaus %
in the absence of inter-agent coordination or centralized control. %

\cref{fig:act_vs_g} shows the robustness of multi-agent active search with TS-UnIK in terms of the number of measurements per agent required by a team to fully recover all OOIs as the number of OOIs $k$ %
increases. For different team sizes ($J$) (indicated as UnIK/TS-UnIK-$J$), %
for both UnIK (R) and TS-UnIK to fully recover all OOIs, the average number of measurements needed per agent in the team increases with more number of OOIs in the search space. %
However in all the settings, TS-UnIK outperforms UnIK (R) %
and the performance gap widens further with more targets $k$ for different $J$. %

\begin{figure}[t]
    \centering
    \begin{subfigure}{0.5\linewidth}
    \centering
    \includegraphics[width=\textwidth]{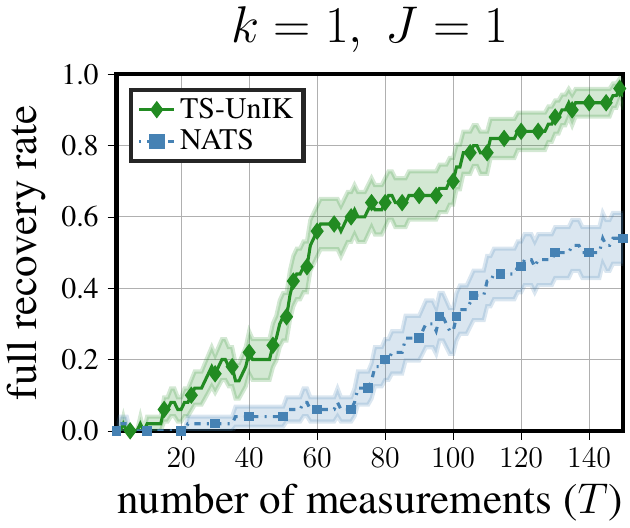}
    \caption{}
    \label{fig:nats_k1}
    \end{subfigure}%
    \begin{subfigure}{0.5\linewidth}
    \centering
    \includegraphics[width=\textwidth]{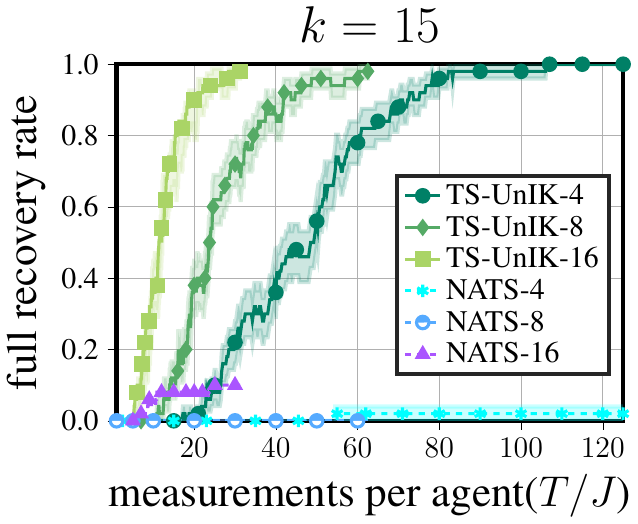}
    \caption{}
    \label{fig:nats_k15}
    \end{subfigure}
    \caption{\textbf{$J$ agents in a $16\times16$ grid with $k$ targets, 50 trials}. TS-based action selection benefits more from the joint uncertainty modeling in UnIK, than with NATS \cite{ghods2021multi}. The performance improvement is particularly significant for multi-agent teams recovering multiple targets in the search space.}
    \label{fig:results_nats}
\end{figure}

\fakeparagraph{TS-UnIK vs. NATS}: 
We now compare the performance of TS-UnIK against NATS with TS based action selection \cite{ghods2021multi}. \cref{fig:nats_k1} shows the full recovery rate within $T = 150$ measurements in a single agent setting ($J = 1$) when there is $k = 1$ target in a $16\times16$ search space. In this setting, TS-UnIK outperforms NATS and enables the single agent to achieve the desired recovery rate more efficiently with significantly fewer measurements. 
\cref{fig:nats_k15} further validates the superiority of TS-UnIK over NATS in a multi-agent setting with multiple OOIs. For different team sizes $J\in\{4,8,16\}$ with the same total number of measurements $T = 500$, we compare the full recovery rate achieved by the team in terms of the number of measurements required per agent. As before, we observe that TS-UnIK is able to efficiently recover all the OOIs and demonstrates a $J$ times improvement in performance as the team size grows by a factor of $J$. On the other hand, with an identical team size and the same number of measurements per agent, NATS fails to achieve full recovery in the majority of trials. The deterioration in performance due to the absence of target location uncertainty modeling in NATS's inference method as observed in \cref{subsec:results_inf}, is further exacerbated in the decentralized multi-agent setting with asynchronous communication. In particular, when different agents perform overlapping sensing actions, they may observe the same OOI at different locations due to observation noise arising from target location uncertainty. As a result, the combined detection and location uncertainty modeling in UnIK not only helps better estimate the possible OOI positions but this improved posterior belief also %
helps TS choose effective sensing actions, leading to %
faster recovery of all OOIs with TS-UnIK.

\subsection{TS-UnIK in Unreal Engine 4 (UE4) with AirSim plugin}
\label{subsec:UE4}

We test TS-UnIK in a pseudo-realistic environment created in UE4 with an AirSim plugin. 
It is a 250m$\times$250m field with trees and animals, discretized into 16$\times$16 grid cells. 
There are $k = 5$ humans randomly positioned within the field, who are the OOIs for our ground robot. 
We use YOLOv3 \cite{redmon2018yolov3} (off-the-shelf) as the agent's object detector, which provides a label and confidence score for OOIs in the agent's FOV (\cref{fig:ue4_yolo}). 
Using the depth maps provided by AirSim, we implement our own depth sensor model that provides noisy location measurements to the agent following \cref{eq:location_uncertainty}. 
Using measurements from these sensors, the agent decides its next sensing action following \cref{eq:reward}, with an additional travel distance penalization term. 
We have included a video demonstrating the performance of TS-UnIK in this environment.\footnote{Also at \href{https://sites.google.com/view/unik-icra23
}{https://sites.google.com/view/
unik-icra23
}.}%
Supporting the results in \cref{subsec:results_actsel}, the agent successfully recovers all OOIs %
with sensing actions that decide where and how the surroundings are observed to eliminate false or missed detections over time.